\documentclass[10pt,final,journal,twocolumn,singlespaced]{IEEEtran}
\usepackage{algorithm}
\usepackage{algpseudocode}
\usepackage{amsmath}
\usepackage{amssymb}
\usepackage{amsthm}
\usepackage{bm}
\usepackage{booktabs}
\usepackage{color}
\usepackage{cite}
\usepackage{epsfig}
\usepackage{epstopdf}
\usepackage{flafter}
\usepackage{float}
\usepackage{ifthen}
\usepackage{multirow}
\usepackage{amsfonts}
\usepackage{makecell} % Make references as [1-4], not [1,2,3,4]
\usepackage{pifont}
\usepackage{subfigure}
\usepackage{times}
\usepackage{threeparttable}
\usepackage{url} % Formatting web addresses % Conditional
\usepackage{subfigure,graphicx}
\usepackage{color,multirow}
\usepackage{makecell}
\usepackage{setspace}
\usepackage{cite} % Make references as [1-4], not [1,2,3,4]
\usepackage{url} % Formatting web addresses
\usepackage{ifthen} % Conditional
\usepackage[T1]{fontenc}
\usepackage{graphicx}
\usepackage{changepage}
\usepackage[colorlinks,linkcolor=red,anchorcolor=gray,citecolor=green,urlcolor=black]{hyperref}
\usepackage{graphics}
\usepackage{booktabs}

\newtheorem{lemma}{Lemma}
\newtheorem{definition}{Definition}
\newtheorem{remark}{Remark}

\newcommand{\tabincell}[2]{\begin{tabular}{@{}#1@{}}#2\end{tabular}}
\usepackage{color}
\definecolor{myGreen}{rgb}{0.945,0.972,0.980}
\begin{document}
\title{H2TF for Hyperspectral Image Denoising: Where Hierarchical Nonlinear Transform Meets Hierarchical Matrix Factorization
%\thanks{(Jiayi Li and Yisi Luo contribute equally to this work.)}
\thanks{The authors are with the School of Mathematical Sciences, University of Electronic Science and Technology of China, Chengdu, China.}}
\author{Jiayi Li,
Jinyu Xie,
Yisi Luo,
Xile Zhao,
Jianli Wang}
\maketitle
\begin{abstract}
Recently, tensor singular value decomposition (t-SVD) has emerged as a promising tool for hyperspectral image (HSI) processing. In the t-SVD, there are two key building blocks: (i) the low-rank enhanced transform and (ii) the accompanying low-rank characterization of transformed frontal slices. Previous t-SVD methods mainly focus on the developments of (i), while neglecting the other important aspect, i.e., the exact characterization of transformed frontal slices. In this letter, we exploit the potentiality in both building blocks by leveraging the \underline{\bf H}ierarchical nonlinear transform and the \underline{\bf H}ierarchical matrix factorization to establish a new \underline{\bf T}ensor \underline{\bf F}actorization (termed as H2TF). Compared to shallow counter partners, e.g., low-rank matrix factorization or its convex surrogates, H2TF can better capture complex structures of transformed frontal slices due to its hierarchical modeling abilities. We then suggest the H2TF-based HSI denoising model and develop an alternating direction method of multipliers-based algorithm to address the resultant model. Extensive experiments validate the superiority of our method over state-of-the-art HSI denoising methods.
\end{abstract}
\IEEEpeerreviewmaketitle
\section{Introduction} \label{sec:Int}
\IEEEPARstart{H}{yperspectral} images (HSIs) inevitably contain mixed noise due to sensor failures or complex imaging conditions \cite{TGRS_stripe,GRSL_He}, which seriously affects subsequent applications. Traditional hand-crafted HSI denoising methods, e.g., low-rankness \cite{GRSL_LR}, total variation (TV) \cite{RCTV}, sparse representations \cite{GRSL_sparse}, and {non-local self-similarity} \cite{NGmeet}, utilize interpretable domain knowledge to design generalizable regularizations for HSI denoising. Their representation abilities may be inferior to data-driven methods using deep neural networks (DNNs) \cite{Hsi-de,sdecnn,HSID-CNN}, which can learn representative denoising mappings via supervised learning with abundant training pairs. However, supervised deep learning methods mostly neglect the prior information of HSIs, which sometimes results in generalization issues over different HSIs and various types of noise.

More recently, tensor singular value decomposition (t-SVD) attracts much attention in HSI denoising \cite{tsvd,SSTVLRTF}. The t-SVD views HSI as an implicit low-rank tensor and exploits the low-rankness in the transformed domain, which can more vividly characterize the structures of HSIs since it is flexible to select {appropriate} transforms and the accompanying low-rank characterization of the transformed {frontal slices}. Under such a framework, there are naturally two {key} building blocks: (i) The selection of the low-rank enhanced transform. A suitable transform can obtain a lower-rank transformed tensor and enhance the recovery quality \cite{TIP_Wang,HLRTF}. (ii) The characterization of low-rankness of transformed frontal slices. The implicit low-rankness of HSIs is exploited by the low-rank modeling of frontal slices in the transformed domain.\par 
	Classical t-SVD-based methods mainly focused on the first building blocks, i.e., the design of different transforms. For example, the discrete Fourier transform (DFT) \cite{DFT} was first used in the t-SVD, and {then} the discrete cosine transform (DCT) \cite{DCT} was employed. Later methods exploited more representative and flexible transforms such as non-invertible transforms \cite{FTNN} and data-dependent transforms \cite{Q-rank} to enhance the low-rankness of {transformed frontal slices}. These methods have achieved increasingly satisfactory results for HSI denoising \cite{tsvd,SSTVLRTF}. Nevertheless, these t-SVD methods pay less attention to the second building block, i.e., the exact characterization of transformed frontal slices. Specifically, they all employ shallow representations such as low-rank matrix factorization (MF) \cite{HLRTF}, QR factorization \cite{qr}, and nuclear norm \cite{FTNN,TIP_Wang} to characterize the transformed frontal slices.\par 
	In this work, we exploit a more representative formulation to capture complex structures of transformed frontal slices. Specifically, we leverage the hierarchical matrix factorization (HMF), which tailors a hierarchical formulation of learnable matrices along with nonlinear layers to capture each frontal slice in the transformed domain. The hierarchical modeling ability of HMF makes it more representative to capture the complex structures of HSIs. Meanwhile, we leverage the hierarchical nonlinear transform (HNT) to enhance the low-rankness of transformed frontal slices. With the \underline{H}ierarchical nonlinear transform and \underline{H}ierarchical matrix factorization, we develop a new \underline{T}ensor \underline{F}actorization method (termed as H2TF) under the t-SVD framework. Correspondingly, we develop the H2TF-based HSI denoising model. {Attributed to the stronger representation abilities of HMF than shallow MF or its surrogates, our H2TF-based model can better capture fine details of the underlying clean HSI than conventional t-SVD-based methods.} Thus, our model is expected to deliver better HSI denoising results. Meanwhile, the parameters of H2TF can be inferred from the observed noisy HSI in an unsupervised manner. In summary, the contributions of this letter are:
	
	{{\bf (i)} We propose a new tensor factorization, i.e., the H2TF, which leverages the expressive power of two key building blocks---the HNT and the HMF, to respectively enhance the low-rankness of transformed data and characterize complex structures of transformed frontal slices. By virtue of their hierarchical modeling abilities, H2TF can faithfully capture fine details of the clean HSI, and thus is beneficial for effectively removing heavy noise in the HSI.}  
	 
	{{\bf (ii)} We suggest an unsupervised H2TF-based HSI denoising model and develop an alternating direction method of multipliers (ADMM)-based algorithm. Extensive experiments on simulated and real-world data validate the superiority of our method over state-of-the-art (SOTA) HSI denoising methods, especially for details preserving and heavy noise removal.}
	\vspace{-0.2cm}
	\section{The Proposed H2TF}
	\subsection{The t-SVD framework}
	We first introduce the general formulation of t-SVD. Suppose that the noisy HSI ${\mathcal Y}\in{\mathbb R}^{h\times w\times b}$ admits ${\mathcal Y} = {\mathcal X} + {\mathcal N}$, where $\mathcal X$ denotes the clean HSI and $\mathcal N$ denotes noise. To infer the underlying clean HSI $\mathcal X$ from the observed $\mathcal Y$, t-SVD method generally formulates the following model:

	\vspace{-0.3cm}
	\begin{small}\begin{equation}
		\label{model_tSVD}
		\begin{split}
			\min_{{\mathcal Z},\theta}\;L({\mathcal Y},{\mathcal X})+\sum_k\psi({\mathcal Z^{(k)}}),\;{\rm where}\;{\mathcal X} = \phi_\theta({\mathcal Z}).
		\end{split}
	\end{equation}\end{small}Here, $L({\mathcal Y},{\mathcal X})$ denotes the fidelity term and $\psi({\mathcal Z}^{(k)})$ represents the low-rank characterization of ${\mathcal Z}^{(k)}$ (which denotes the $k$-th frontal (spatial) slice of ${\mathcal Z}\in{\mathbb R}^{h\times w\times b}$ \cite{FTNN}). $\phi_\theta(\cdot):{\mathbb R}^{h\times w\times b}\rightarrow{\mathbb R}^{h\times w\times b}$ denotes a transform with learnable parameters $\theta$, which transforms the low-rank representation $\mathcal Z$ into the original domain. Sometimes the transform $\phi_\theta(\cdot)$ may not be learnable (e.g., the fixed DFT \cite{DFT}), and in those situations the optimization variable only includes $\mathcal Z$. \par 
	The philosophy of the t-SVD model (\ref{model_tSVD}) is to minimize the rank in the transformed domain, which can model the implicit low-rankness of HSI. There are naturally two key building blocks for exactly modeling the implicit low-rankness, i.e., the selection of the transform $\phi_\theta(\cdot)$ and the exact low-rank characterization $\psi(\cdot)$ of the transformed frontal slice ${\mathcal Z}^{(k)}$. Most t-SVD-based methods focus on the design of different transforms $\phi_\theta(\cdot)$ (see examples in \cite{FTNN,Q-rank,HLRTF}), but all of them pay less attention to the exact characterization of the transformed frontal slice. {They mostly adopt shallow representations to characterize ${\mathcal Z}^{(k)}$, e.g., MF \cite{HLRTF,Pan_TIP}, QR factorization \cite{qr}, and nuclear norm \cite{DCT,FTNN}.} However, {these shallow representations may not be expressive enough to capture fine details of the clean HSI.} Therefore, more representative methods are {desired} to enhance the representation abilities of the model in the transformed domain.
	\vspace{-0.2cm}
	\subsection{HMF for Characterizing ${\mathcal Z}^{(k)}$}
	To cope with this challenge, we leverage the HMF (hierarchical matrix factorization) to characterize ${\mathcal Z}^{(k)}$. The hierarchical modeling ability of HMF helps it more faithfully capture complex structures of the transformed frontal slice ${\mathcal Z}^{(k)}$ than shallow counter partners, e.g., SVD, MF, and QR factorization.\par 
	The standard MF used in previous t-SVD methods \cite{Pan_TIP,HLRTF} decomposes a low-rank matrix $\mathbf{Z} \in \mathbb{R}^{h\times w}$ into two factors as $\mathbf{Z} = \mathbf{W}_2\mathbf{W}_1$, where ${\bf W}_2\in{\mathbb R}^{h\times r}$, ${\bf W}_1\in{\mathbb R}^{r\times w}$, and $r$ is the rank. To model the hierarchical structures of $\bf Z$, we extend the MF to the product of multiple matrix factors $\{\mathbf{W}_d\}_{d=1}^l$:

\vspace{-0.25cm}
	\begin{small}\begin{equation}\label{HMF_linear}
		\mathbf{Z} = \mathbf{W}_l{\bf W}_{l-1}\cdots\mathbf{W}_1,\vspace{-0.1cm}
	\end{equation}\end{small}where $\mathbf{W}_d \in \mathbb{R}^{r_{d} \times r_{d-1}}$, $ r_{l} = h$, and $r_{0} = w$. It was shown in \cite{arora2019implicit} that such a linear HMF can induce an implicit low-rank regularization on $\bf Z$ when using gradient-based optimization. Generally, the larger $l$ is (i.e., adding depth to the HMF), the tendency towards low-rank solutions goes stronger and oftentimes leads to better matrix recovery performances. Thus, the HMF is suitable to play the role of low-rank regularization in the t-SVD model (\ref{model_tSVD}).\par
	Nevertheless, the linear HMF (\ref{HMF_linear}) may not be sufficient to capture nonlinear interactions inside HSIs. It motivates us to utilize the nonlinear HMF \cite{air-net,NN_DMF} to model the low-rank matrix $\bf Z$ via $
		\mathbf{Z} = \mathbf{W}_{l}\sigma(\mathbf{W}_{l-1}\cdots \mathbf{W}_{3}\sigma(\mathbf{W}_{2}\mathbf{W}_{1}))
		\label{equ2}$, where $\sigma(\cdot)$ is a nonlinear scalar function. Classical HMF-based methods \cite{arora2019implicit,air-net} only utilize HMF to tackle the two-dimensional matrix. However, matrixing the HSI inevitably destroys its high-dimensional data {structures}. Therefore, we suggest tailoring $b$ nonlinear HMFs to model the transformed tensor $\mathcal Z$ by using each HMF to represent one of the frontal slices of $\mathcal Z$. Formally, we represent each frontal slice of $\mathcal Z$ by
		
		\vspace{-0.3cm}
	\begin{small}\begin{equation*}
		\begin{split}
			{\mathcal Z}^{(k)} =  \mathcal{W}^{(k)}_{l}\sigma(\mathcal{W}^{(k)}_{l-1}\cdots \mathcal{W}^{(k)}_{3}\sigma&(\mathcal{W}^{(k)}_{2}\mathcal{W}^{(k)}_{1})),k=1,2,\cdots,b.
		\end{split}
	\end{equation*}\end{small}The above HMFs can be equivalently formulated as the tensor formulation $
		\mathcal{Z} = {\mathcal W}_l\Delta\sigma({\mathcal W}_{l-1}\Delta\cdots{\mathcal W}_3\Delta\sigma({\mathcal W}_2\Delta{\mathcal W}_1))$, where $\Delta$ is the tensor face-wise product \cite{kernfeld2015tensor} and $\{\mathcal{W}_d \in \mathbb{R}^{r_{d} \times r_{d-1}\times b}\}_{d=1}^l$ are some factor tensors.\par
	Compared to shallow counter partners, e.g., MF, QR factorization, and nuclear norm, the above nonlinear HMF can better capture complex hierarchical structures of HSIs due to its nonlinear hierarchical modeling abilities, which helps to better recover fine details of HSI and remove heavy noise.
	\begin{figure}[t]
		\centering
		\includegraphics[width=0.45\textwidth]{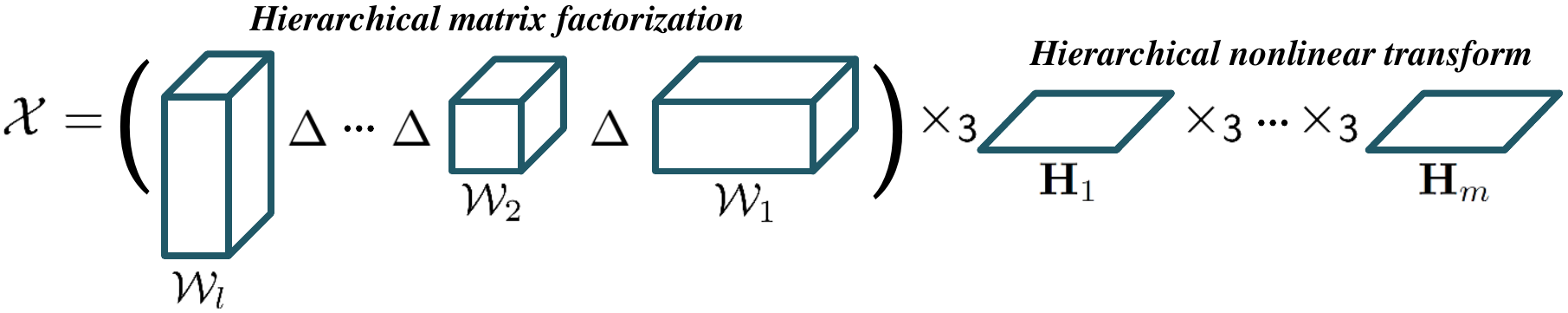}
		\vspace{-0.4cm}
		\caption{A general illustration of the H2TF representation of a tensor $\mathcal X$. The nonlinear layer $\sigma(\cdot)$ is omitted for space consideration.}
		\label{fig:H2TF}
		\vspace{-0.6cm}
	\end{figure}
		\vspace{-0.2cm}
	\subsection{The Proposed H2TF}
	Next, we {introduce} our H2TF. {Recall} that two {key} building blocks in the t-SVD are the selection of the transform $\phi_\theta(\cdot)$ and the characterization of the transformed frontal slice ${\mathcal Z}^{(k)}$. We have leveraged the HMF to characterize ${\mathcal Z}^{(k)}$, and we further leverage the HNT (hierarchical nonlinear transform) as the first building block $\phi_\theta(\cdot)$:
	
	\vspace{-0.4cm}
	\begin{small}\begin{equation*}
		\phi_\theta({\mathcal Z}) := \sigma(\cdots\sigma({\mathcal Z}\times_3 {\bf H}_1)\times_3\cdots\times_3{\bf H}_{m-1})\times_3 {\bf H}_{m},
		\label{equ:7}
	\end{equation*}\end{small}where $\sigma(\cdot)$ is a nonlinear scalar function, $\theta:=\{{\bf H}_p\in{\mathbb R}^{b\times b}\}_{p=1}^m$
	are learnable parameters of HNT, and $\times_3$ is the mode-3 tensor-matrix product \cite{SIAM_13}. It was throughout demonstrated \cite{HLRTF} that the HNT can effectively enhance the low-rankness of transformed tensor and thus obtain a better low-rank representation than shallow transforms (e.g., DFT \cite{DFT} and DCT \cite{DCT}), which benefits the implicit low-rank modeling.
	\begin{definition}[H2TF]
		Finally, we can define the following factorization modality of a certain low-rank tensor ${\mathcal X}$ parameterized by $\{{\mathcal W}_d\}_{d=1}^l$ and $\{{\bf H}_p\}_{p=1}^m$:
		
		\vspace{-0.4cm}
		\begin{small}\begin{equation}\label{H2TF}
			\begin{split}
				&{\mathcal X} = \phi_\theta\big{(}\underbrace{{\mathcal W}_l\Delta\sigma({\mathcal W}_{l-1}\Delta\cdots{\mathcal W}_3\Delta\sigma({\mathcal W}_2\Delta{\mathcal W}_1))}_{\rm Hierarchical\;matrix\;factorization}\big{)},\\
				&\quad\phi_\theta({\mathcal Z}):=\underbrace{\sigma(\cdots\sigma({\mathcal Z}\times_3 {\bf H}_1)\times_3\cdots\times_3{\bf H}_{m-1})\times_3 {\bf H}_{m}}_{\rm Hierarchical\;nonlinear\;transform},
			\end{split}
		\end{equation}\end{small}which we call the H2TF representation of ${\mathcal X}$.  
	\end{definition}
	A general illustration of the proposed H2TF is shown in Fig. \ref{fig:H2TF}. H2TF benefits from the HMF to exploit complex hierarchical information of transformed frontal slices and the HNT to enhance the low-rankness in the transformed domain. With the hierarchical modeling abilities of H2TF, the characterization of HSIs would be more accurate. Therefore, H2TF can more faithfully capture fine details and rich textures of HSIs and remove heavy mixed noise. Now, we discuss the connections between H2TF and some popular matrix/tensor factorizations.
	\begin{remark}
		By changing the layer number of hierarchical matrix factorization (i.e., $l$) and the layer number of hierarchical nonlinear transform (i.e., $m$), H2TF includes many matrix/tensor factorizations as special cases:\par {\bf (i)} When $l=2$, i.e., the HMF degenerates into the MF, our H2TF degenerates into the hierarchical low-rank tensor factorization \cite{HLRTF}. {\bf (ii)} When $m=1$ and ${\bf H}_m$ is an identity matrix (i.e., the transform $\phi_\theta(\cdot)$ is an identical mapping), our H2TF degenerates into the {plain} HMFs \cite{air-net,NN_DMF} applied on each frontal slice of the tensor separately. In the following, we {interpret} this case as ``$m=0$'' since the transform is {neglected}. {\bf (iii)} When $l=2$ and $m=1$ with ${\bf H}_m$ being the fixed inverse DFT matrix, our H2TF degenerates into the classical low-tubal-rank tensor factorization \cite{liu2019low,Pan_TIP}. 
	\end{remark}
	Moreover, H2TF can explicitly preserve the low-rankness of the tensor when omitting some nonlinearity, as stated below.
	{\begin{lemma}\label{lemma_1}
			Suppose that ${\mathcal X} = \phi\big{(}{\mathcal W}_l\Delta({\mathcal W}_{l-1}\Delta\cdots\Delta{\mathcal W}_1)\big{)}\in{\mathbb R}^{h\times w\times b}$, where $\{\mathcal{W}_d \in \mathbb{R}^{r_{d} \times r_{d-1}\times b}\}_{d=1}^l$ ($ r_{l} = h$ and $r_{0} = w$) are factor tensors, $\phi({\mathcal Z}):={\mathcal Z}\times_3 {\bf F}^{-1}$ is the inverse DFT, and ${\bf F}^{-1}$ is the inverse DFT matrix (which is a special case of H2TF). Then we have ${\rm rank}_t({\mathcal X})\leq\min\{r_0,r_1,\cdots,r_l\}$, where ${\rm rank}_t(\cdot)$ denotes the tensor tubal-rank \cite{DFT,TIP_Wang,HLRTF}.
	\end{lemma}}
	Lemma \ref{lemma_1} indicates that H2TF can preserve the low-rankness in the linear special case, where the degree of low-rankness (the upper bound of tubal-rank) is conditioned on the sizes of factor tensors. Therefore, we can readily control the degree of low-rankness by tuning the sizes of factor tensors in {H2TF}.
		\vspace{-0.2cm}
	\subsection{H2TF for HSI Denoising}
	{H2TF is a potential tool for multi-dimensional data {analysis and processing}. We consider HSI denoising as a representative real-world application. By applying the H2TF representation (\ref{H2TF}) into (\ref{model_tSVD}), we can obtain the following HSI denoising model:}

\vspace{-0.4cm}
{\begin{small}\begin{equation*}
		\begin{split}
			&\min_{\{{\mathcal W}_d\}_{d=1}^l,\{{\bf H}_p\}_{p=1}^m}\;L({\mathcal Y},{\mathcal X}),\\
			&\;\;{\rm where}\;{\mathcal X} = \phi_\theta\big{(}{\mathcal W}_l\Delta\sigma({\mathcal W}_{l-1}\Delta\cdots{\mathcal W}_3\Delta\sigma({\mathcal W}_2\Delta{\mathcal W}_1))\big{)}.
		\end{split}
	\end{equation*}\end{small}}In the HSI denoising problem, we consider the fidelity term as $L({\mathcal Y},{\mathcal X}) = \lVert{\mathcal Y}-{\mathcal X}-{\mathcal S} \rVert_F^2 + \alpha_1\lVert{\mathcal S}\rVert_{\ell_1}$, where $\lVert\cdot \rVert_F^2$ denotes the Frobenius norm and we introduce ${\mathcal S}\in{\mathbb R}^{h\times w\times b}$ to represent sparse noise (often contains impulse noise and stripes). The $\ell_1$-norm enforces the sparsity on $\mathcal{S}$ so that the sparse noise can be eliminated. Here, $\alpha_1$ is a trade-off parameter.\par
	Meanwhile, our H2TF can be readily combined with other proven techniques to enhance the denoising abilities. Here, we consider the hybrid spatial-spectral TV (HSSTV) regularization \cite{SS_TV_ICIP} to further capture spatial and spatial-spectral local smoothness of HSIs. The HSSTV is formulated as $\lVert\mathcal{X}\rVert_{\rm HSSTV} := \alpha_2\lVert\mathcal{X}\rVert_{\rm TV}+\alpha_3\lVert\mathcal{X}\rVert_{\rm SSTV}$, where $\lVert\mathcal{X}\rVert_{\rm TV} := \lVert\nabla_x \mathcal{X}\rVert_{\ell_1} + \lVert\nabla_y \mathcal{X}\rVert_{\ell_1}$, $\lVert\mathcal{X}\rVert_{\rm SSTV} := \lVert\nabla_x (\nabla_z\mathcal{X})\rVert_{\ell_1} + \lVert\nabla_y (\nabla_z\mathcal{X})\rVert_{\ell_1}$, and $\alpha_i$ ($i=2,3$) are trade-off parameters. Here, the derivative operators are defined as $(\nabla_x\mathcal{X})_{(i,j,k)} := \mathcal{X}_{(i+1,j,k)}-\mathcal{X}_{(i,j,k)}$, $(\nabla_y\mathcal{X})_{(i,j,k)} := \mathcal{X}_{(i,j+1,k)}-\mathcal{X}_{(i,j,k)}$, and $(\nabla_z\mathcal{X})_{(i,j,k)} := \mathcal{X}_{(i,j,k+1)}-\mathcal{X}_{(i,j,k)}$, where $\mathcal{X}_{(i,j,k)}$ denotes the $(i,j,k)$-th element of $\mathcal{X}$.\par
	Based on the formulations of fidelity term and HSSTV, the proposed H2TF-based HSI denosing model {is formulated as}
	
	\vspace{-0.4cm}
	\begin{small}\begin{equation}\label{model_final}
		\begin{split}
			&\min_{\{{\mathcal W}_d\}_{d=1}^l,\{{\bf H}_p\}_{p=1}^m,{\mathcal S}}\;\lVert{\mathcal Y}-{\mathcal X}-{\mathcal S} \rVert_F^2 +\alpha_1\lVert{\mathcal S}\rVert_{\ell_1}+\lVert{\mathcal X}\rVert_{\rm HSSTV},\\
			&\quad\quad{\rm where}\;{\mathcal X} = \phi_\theta\big{(}{\mathcal W}_l\Delta\sigma({\mathcal W}_{l-1}\Delta\cdots{\mathcal W}_3\Delta\sigma({\mathcal W}_2\Delta{\mathcal W}_1))\big{)}.
		\end{split}
	\end{equation}\end{small}\par 
	Compared to previous t-SVD-based HSI denoising methods \cite{tsvd,SSTVLRTF}, H2TF has powerful representation abilities brought from the hierarchical structures and thus could better capture fine details of HSIs. Besides, the parameters of H2TF are unsupervisedly inferred from the noisy HSI by optimizing (\ref{model_final}) without the requirement of training process.
		\vspace{-0.2cm}
	\subsection{ADMM-Based Algorithm}
	To tackle the problem (\ref{model_final}), we {develop} an ADMM-based algorithm. By introducing auxiliary variables $\mathcal{V}_i$ ($i =1, 2, 3, 4$), (\ref{model_final}) can be equivalently formulated as
	
	\vspace{-0.4cm}
	\begin{small}\begin{equation*}
		\begin{split}
			&\min_{\substack{\{{\mathcal W}_d\}_{d=1}^l,\{{\bf H}_p\}_{p=1}^m,\\
					{\mathcal S},\{{\mathcal V}_i\}_{i=1}^4}} 
			\lVert\mathcal{Y}-\mathcal{X}-\mathcal{S}\lVert^2_F + \alpha_1\lVert\mathcal{S}\rVert_{\ell_1}+\alpha_2\lVert\mathcal{V}_1\rVert_{\ell_1}+\\
			&\quad\quad\quad\quad\quad\quad\quad\alpha_2\lVert\mathcal{V}_2\rVert_{\ell_1} +\alpha_3\lVert\mathcal{V}_3\rVert_{\ell_1}+\alpha_3\lVert\mathcal{V}_4\rVert_{\ell_1} ,\\
			&{\rm s.t.}\;\;\mathcal{V}_1=\nabla_x\mathcal{X}, \mathcal{V}_2=\nabla_y\mathcal{X},\quad\mathcal{V}_3=\nabla_x(\nabla_z\mathcal{X}), \mathcal{V}_4=\nabla_y(\nabla_z\mathcal{X}),
		\end{split}
	\end{equation*}\end{small}where ${\mathcal X} = \phi_\theta\big{(}{\mathcal W}_l\Delta\sigma({\mathcal W}_{l-1}\Delta\cdots{\mathcal W}_3\Delta\sigma({\mathcal W}_2\Delta{\mathcal W}_1))\big{)}$. The corresponding augmented Lagrangian function is 
	
	\vspace{-0.4cm}
	\begin{small}\begin{equation*}
		\begin{split}
			&\mathcal{L}_\mu(\{{\mathcal W}_d\}_{d=1}^l, \{{\bf H}_p\}_{p=1}^m, \mathcal{S}, \{\mathcal{V}_i\}_{i=1}^4, \{\varLambda_i\}_{i=1}^4) \\
			=&\;\lVert\mathcal{Y}-\mathcal{X}-\mathcal{S}\lVert^2_F + \alpha_1\lVert\mathcal{S}\rVert_{\ell_1}+ 
			\alpha_2\lVert\mathcal{V}_1\rVert_{\ell_1}+\alpha_2\lVert\mathcal{V}_2\rVert_{\ell_1}+ \\
			&\alpha_3\lVert\mathcal{V}_3\rVert_{\ell_1} 
			+\alpha_3\lVert\mathcal{V}_4\rVert_{\ell_1}
			+ \frac \mu 2\lVert\nabla_x\mathcal{X}+\frac{\varLambda_1}{\mu}-\mathcal{V}_1\rVert_F^2 + \\
			& \frac \mu 2\lVert\nabla_y\mathcal{X}+\frac{\varLambda_2}{\mu}-\mathcal{V}_2\rVert_F^2  + \frac \mu 2\lVert\nabla_x(\nabla_z\mathcal{X})+\frac{\varLambda_3}{\mu}-\mathcal{V}_3\rVert_F^2 + \\
			&\frac \mu 2\lVert\nabla_y(\nabla_z\mathcal{X})+\frac{\varLambda_4}{\mu}-\mathcal{V}_4\rVert_F^2,       
		\end{split}
	\end{equation*}\end{small}where $\mu$ is the penalty parameter, $\varLambda_i$ $(i =1, 2, 3, 4)$ are multipliers, and $\mathcal X$ is defined as in (\ref{H2TF}). The joint minimization problem can be decomposed into easier subproblems, followed by the update of Lagrangian multipliers.
	
	The $\mathcal{V}_i$ $(i=1,2,3,4)$ subproblems are
	
	\vspace{-0.4cm}
	\begin{small}\begin{equation*}
		\begin{cases}
			\min_{\mathcal{V}_1}\frac \mu 2\lVert\nabla_x\mathcal{X}^t+\frac{\varLambda_1^t}{\mu}-\mathcal{V}_1\rVert_F^2 + \alpha_2\lVert\mathcal{V}_1\rVert_{\ell_1}\\
			\min_{\mathcal{V}_2}\frac \mu 2\lVert\nabla_y\mathcal{X}^t+\frac{\varLambda_2^t}{\mu}-\mathcal{V}_2\rVert_F^2 + \alpha_2\lVert\mathcal{V}_2\rVert_{\ell_1}\\
			\min_{\mathcal{V}_3}\frac \mu 2\lVert\nabla_x(\nabla_z\mathcal{X}^t)+\frac{\varLambda_3^t}{\mu}-\mathcal{V}_3\rVert_F^2+ \alpha_3\lVert\mathcal{V}_3\rVert_{\ell_1}\\
			\min_{\mathcal{V}_4}\frac \mu 2\lVert\nabla_y(\nabla_z\mathcal{X}^t)+\frac{\varLambda_4^t}{\mu}-\mathcal{V}_4\rVert_F^2+ \alpha_3\lVert\mathcal{V}_4\rVert_{\ell_1},
		\end{cases}
	\end{equation*}\end{small}which can be exactly solved by $\mathcal{V}^{t+1}_1 = Soft_{\frac{\alpha_2}{\mu}}(\nabla_x\mathcal{X}^t+\frac{\varLambda_1^t}{\mu})$, $\mathcal{V}^{t+1}_2 = Soft_{\frac{\alpha_2}{\mu}}(\nabla_y\mathcal{X}^t+\frac{\varLambda_2^t}{\mu})$, $\mathcal{V}^{t+1}_3 = Soft_{\frac{\alpha_3}{\mu}}(\nabla_x(\nabla_z\mathcal{X}^t)+\frac{\varLambda_3^t}{\mu})$, and $\mathcal{V}^{t+1}_4 = Soft_{\frac{\alpha_3}{\mu}}(\nabla_y(\nabla_z\mathcal{X}^t)+\frac{\varLambda_4^t}{\mu})$, where $\big{(}Soft_{v}({\mathcal X})\big{)}_{(i,j,k)}:=\text{sign}({\mathcal X}_{(i,j,k)})\max\{|{\mathcal X}_{(i,j,k)}|-v, 0\}$.
	
	The $\mathcal{S}$ subproblem is $	\min_{\mathcal{S}} \lVert\mathcal{Y}-\mathcal{X}^t-\mathcal{S}\lVert^2_F + \alpha_1\lVert\mathcal{S}\rVert_{\ell_1}$, which can be exactly solved by $\mathcal{S}^{t+1} = Soft_{\frac{\alpha_1}{2}}(\mathcal{Y}-\mathcal{X}^t)$.
	
   The $\mathcal{X}$ subproblem is
   
   \vspace{-0.4cm}
	\begin{small}\begin{equation*}
		\begin{split}
			&\min_{\{{\mathcal W}_d\}_{d=1}^l,\{{\bf H}_p\}_{p=1}^m}\; \lVert\mathcal{Y}-\mathcal{X}-\mathcal{S}^t\lVert^2_F 
			+ \frac\mu2(\lVert\nabla_x\mathcal{X}-\mathcal{D}_1^t\rVert_F^2 +\\ 
			&\lVert\nabla_y\mathcal{X}-\mathcal{D}_2^t\rVert_F^2     +  \lVert\nabla_x(\nabla_z\mathcal{X})-\mathcal{D}_3^t\rVert_F^2 +\lVert\nabla_y(\nabla_z\mathcal{X})-\mathcal{D}_4^t\rVert_F^2),           
		\end{split}
	\end{equation*}\end{small}where $\mathcal{D}^t_i := \mathcal{V}_i^{t} -\frac{\varLambda_i^t} {\mu}$ ($i=1,2,3,4$). {The clean HSI} $\mathcal{X}$ is parameterized by $\{{\mathcal W}_d\}_{d=1}^l$ and $\{{\bf H}_p\}_{p=1}^m$, as presented in Eq. (\ref{H2TF}). To tackle the nonlinear and nonconvex $\mathcal{X}$ subproblem, we apply the adaptive moment estimation (Adam) algorithm \cite{Adam}. In each iteration of the ADMM-based algorithm, we employ one step of the Adam to update $\{{\mathcal W}_d\}_{d=1}^l$ and $\{{\bf H}_p\}_{p=1}^m$.
	
	Finally, the Lagrange multipliers are updated by $\varLambda_1^{t+1} = \varLambda_1^{t} + \mu(\nabla_x\mathcal{X}^t-\mathcal{V}_1^{t})$, $\varLambda_2^{t+1} = \varLambda_2^{t} + \mu(\nabla_y\mathcal{X}^t-\mathcal{V}_2^{t})$, $\varLambda_3^{t+1} = \varLambda_3^{t} + \mu(\nabla_x(\nabla_z\mathcal{X}^t)-\mathcal{V}_3^{t})$, and $\varLambda_4^{t+1} = \varLambda_4^{t} + \mu(\nabla_y(\nabla_z\mathcal{X}^t)-\mathcal{V}_4^{t})$.
	\begin{figure*}[t]
		\tiny
		\setlength{\tabcolsep}{0.9pt}
		\begin{center}
			\begin{tabular}{ccccccccc}
				\includegraphics [width=0.09\textwidth]{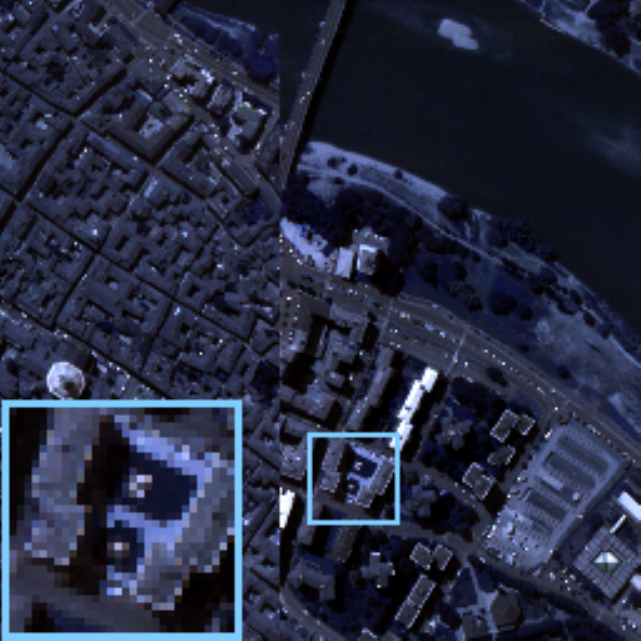}&
                \includegraphics [width=0.09\textwidth]{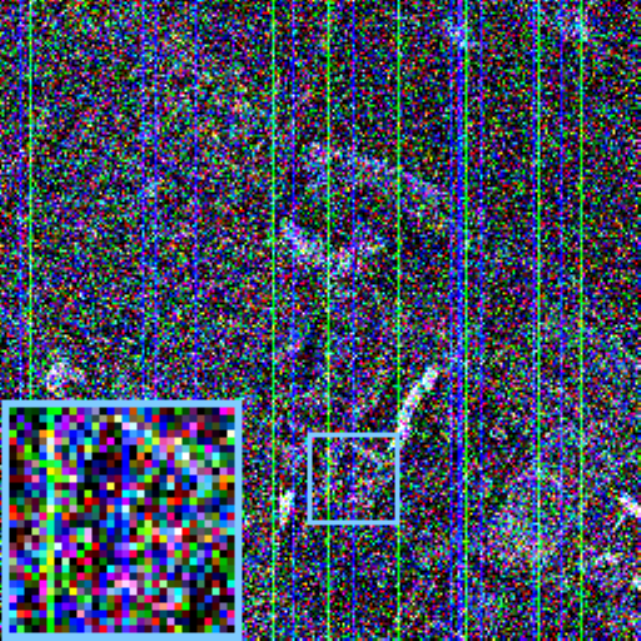}&
				\includegraphics [width=0.09\textwidth]{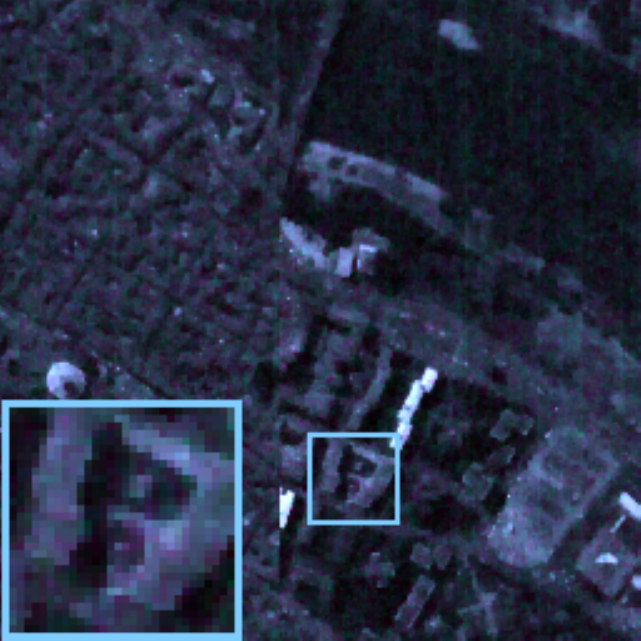}&
				\includegraphics [width=0.09\textwidth]{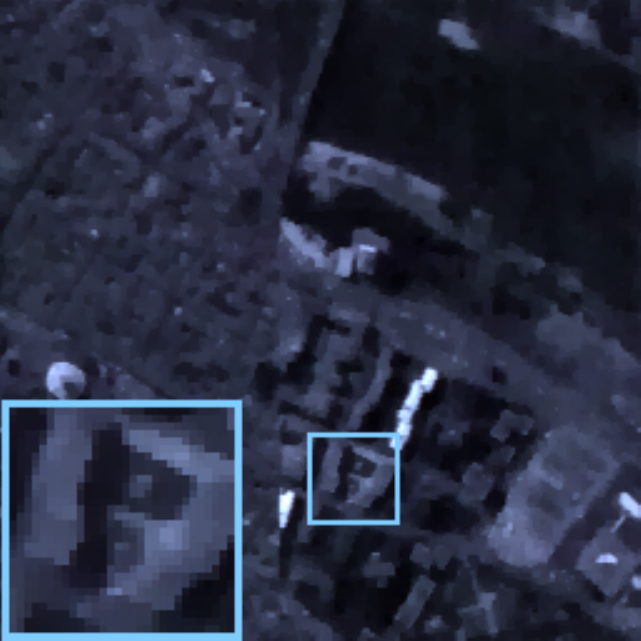}&
				\includegraphics [width=0.09\textwidth]{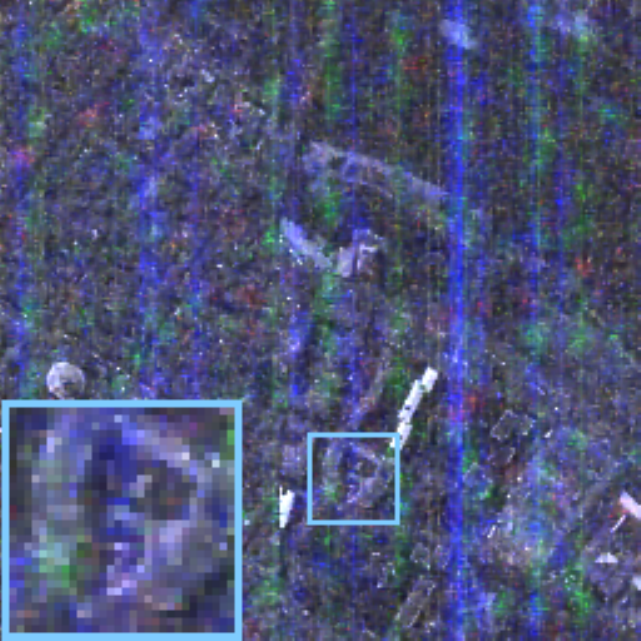}&
				\includegraphics [width=0.09\textwidth]{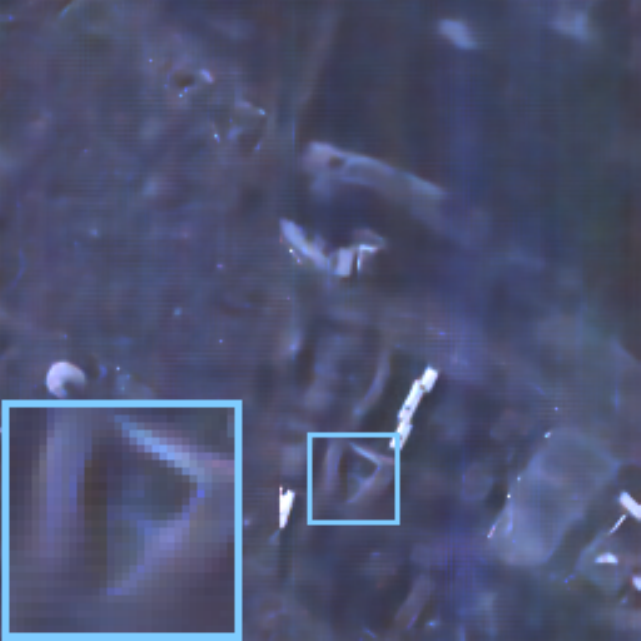}&
				\includegraphics [width=0.09\textwidth]{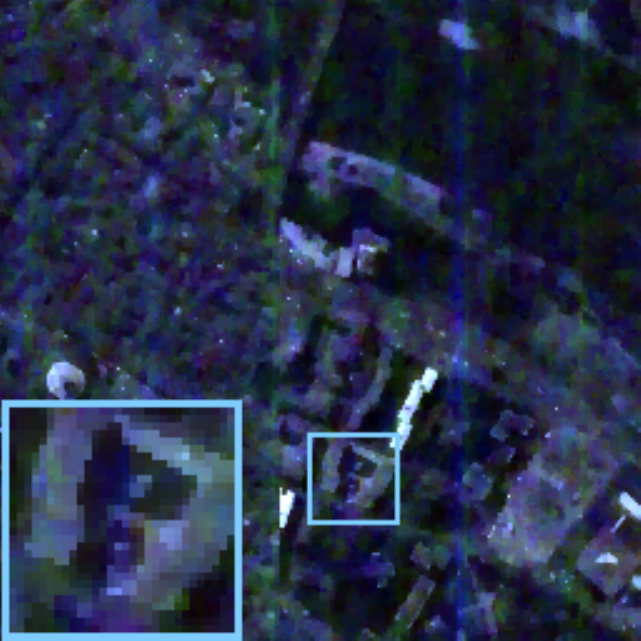}&
				\includegraphics [width=0.09\textwidth]{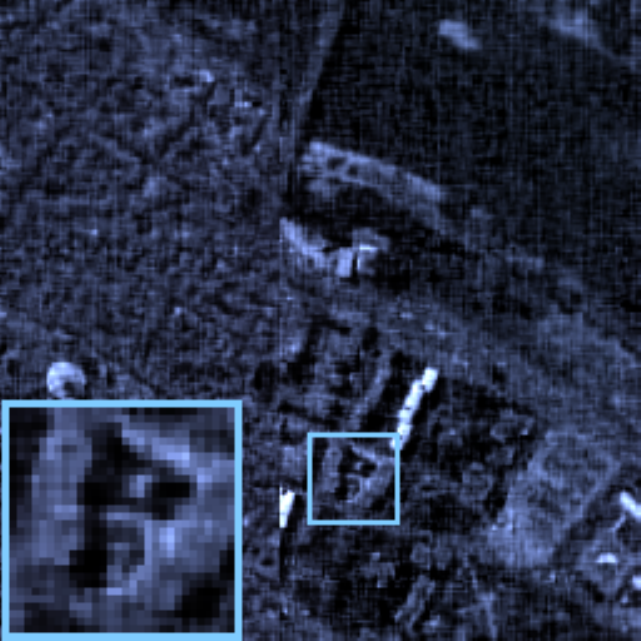}&
				\includegraphics [width=0.09\textwidth]{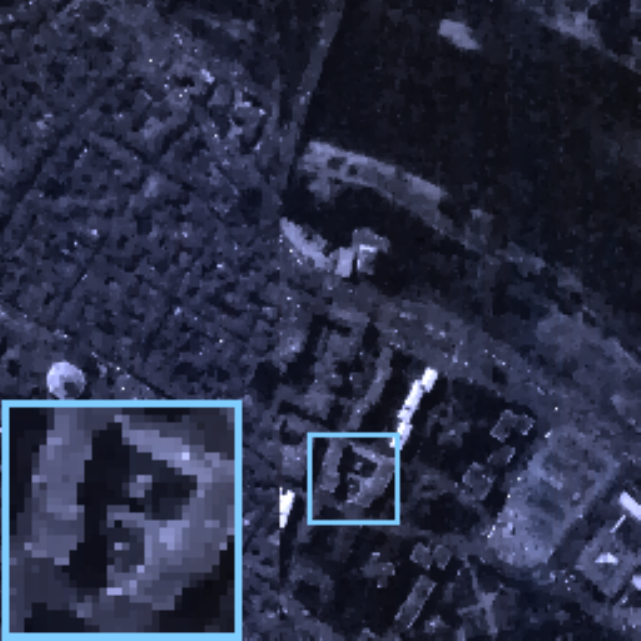}\\
				\includegraphics [width=0.09\textwidth]{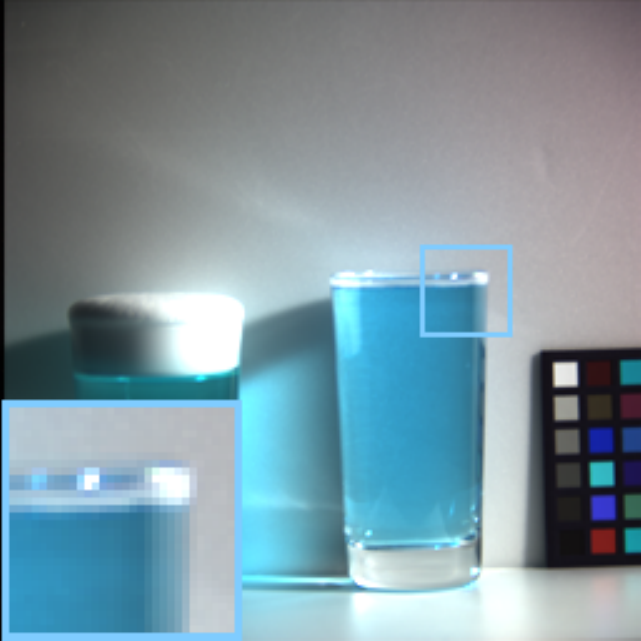}&
				\includegraphics [width=0.09\textwidth]{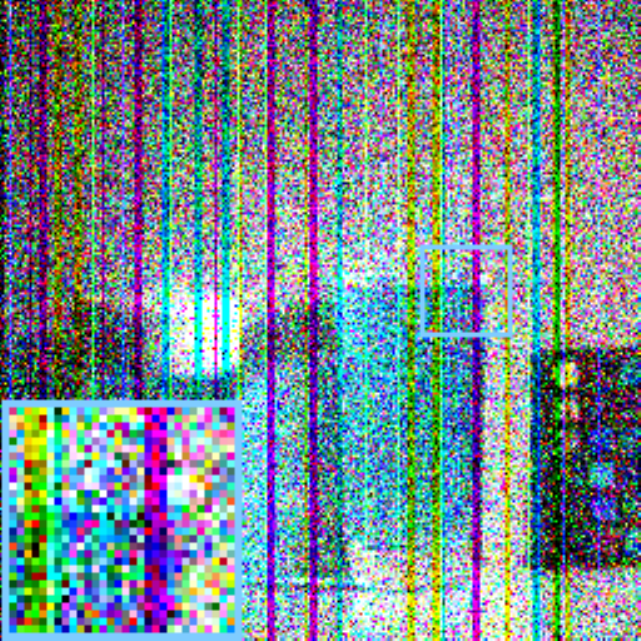}&
				\includegraphics [width=0.09\textwidth]{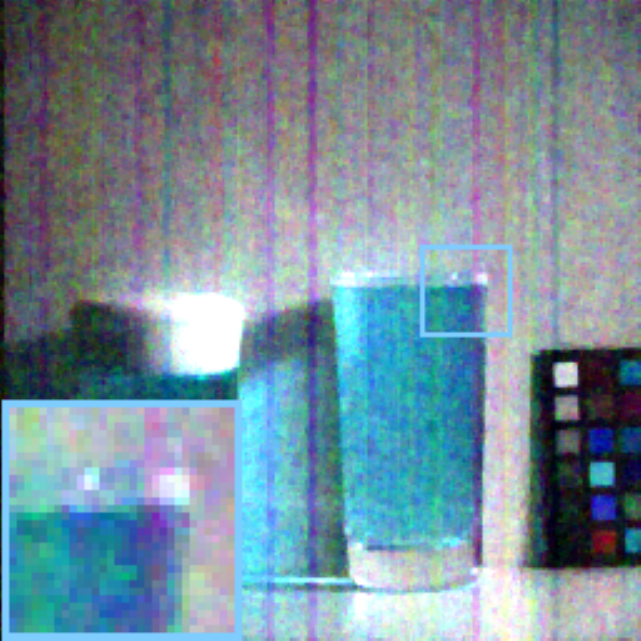}&
				\includegraphics [width=0.09\textwidth]{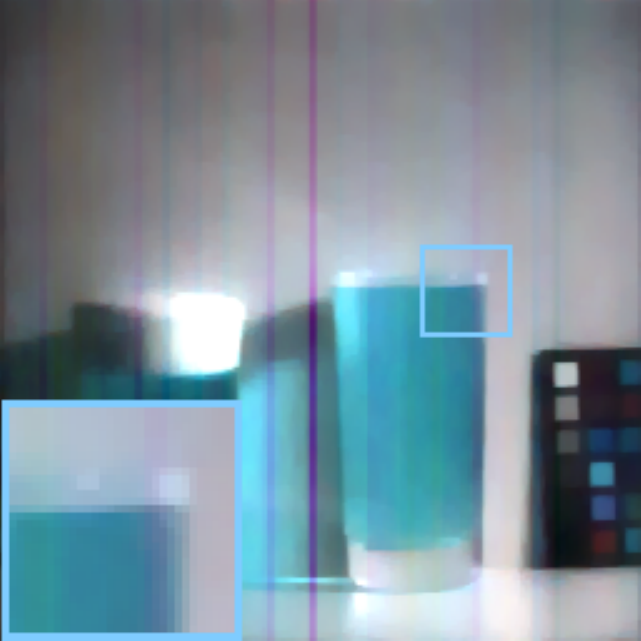}&
				\includegraphics [width=0.09\textwidth]{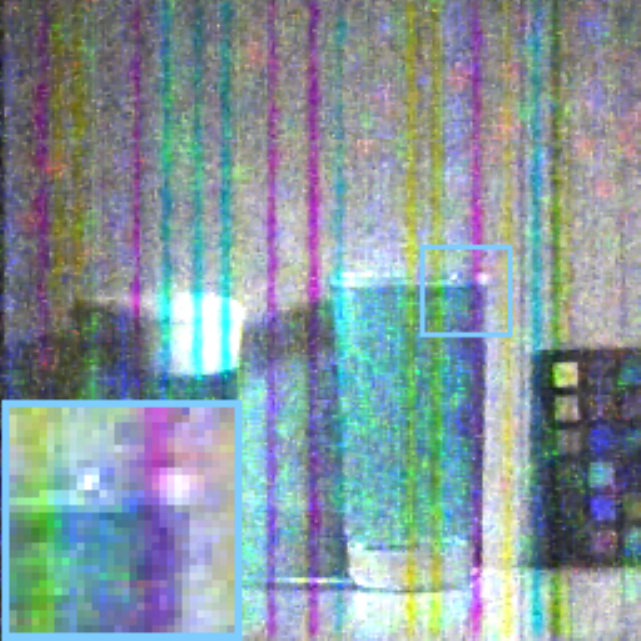}&
				\includegraphics[width=0.09\textwidth]{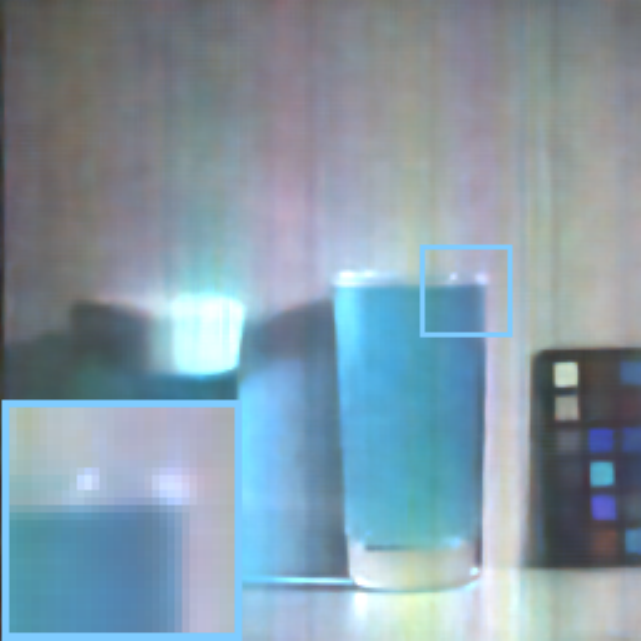}&
				\includegraphics [width=0.09\textwidth]{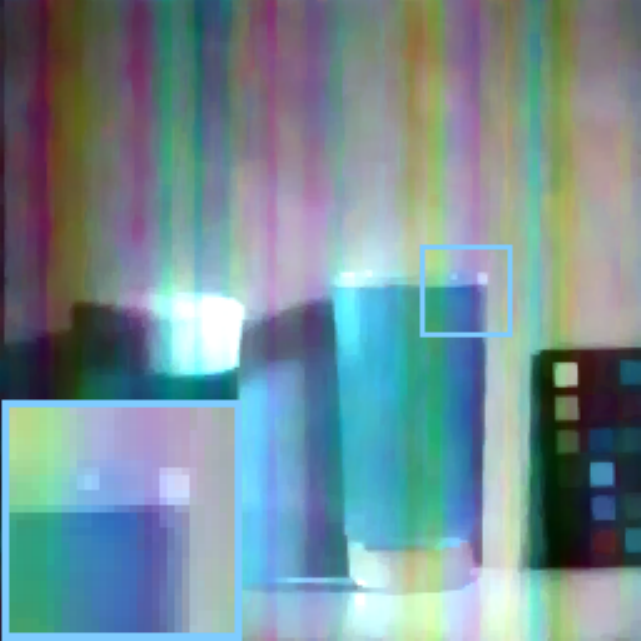}&
				\includegraphics [width=0.09\textwidth]{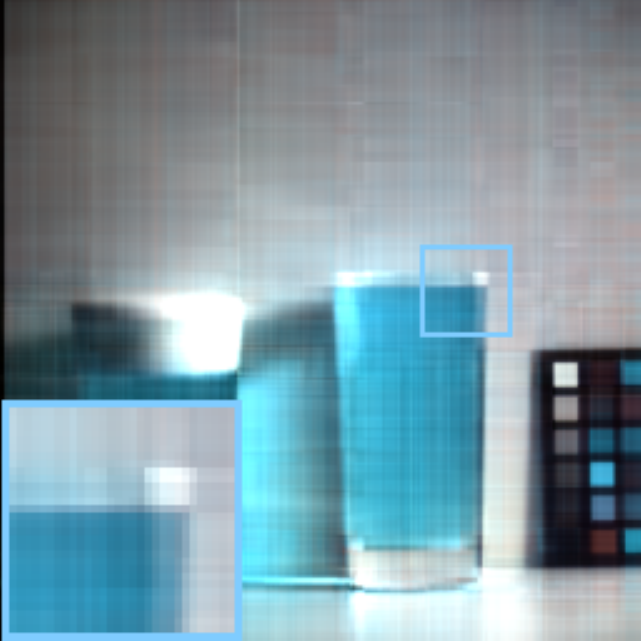}&
				\includegraphics [width=0.09\textwidth]{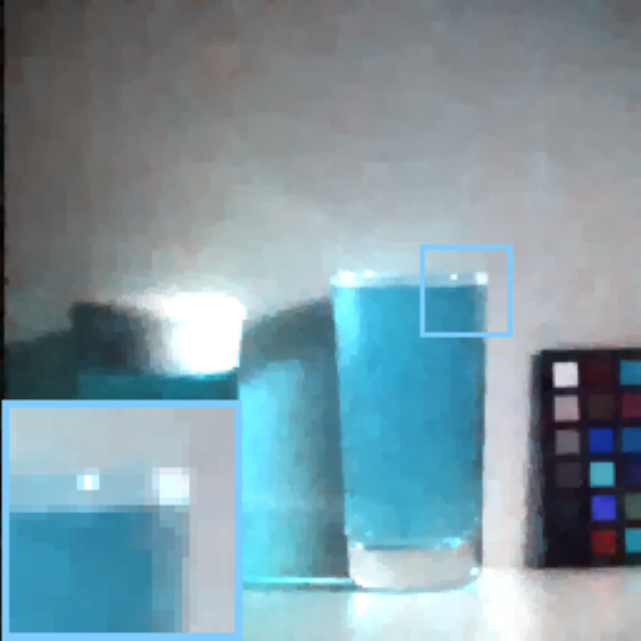}\\
				\vspace{-0.4cm}
		Original &Observed & LRTDTV & SSTV-LRTF & HSID-CNN & SDeCNN &RCTV& HLRTF & H2TF\\
		\end{tabular}
	\end{center}
		\caption{Pseudo-color images of HSI denoising results by different methods on simulated data {\it PaviaC} Case 4 (first row) and {\it Cups} Case 5 (second row). }
		\label{vis}
		\vspace{-0.3cm}
	\end{figure*}
 \begin{table}[t!]
 	\setlength\tabcolsep{2.3pt}
 	\tiny
 	% \scriptsize
 	\centering
 	\caption{Average quantitative denoising results by different methods.\vspace{-0.2cm}}
 	\begin{spacing}{0.8}
 		\scalebox{1.1}{%
 			\begin{tabular}{clrlrlrlrlrl}
 				\toprule
 				\multirow{3}{*}{Dataset} &\multirow{3}{*}{Method} & \multicolumn{2}{c}{Case 1} & \multicolumn{2}{c}{Case 2}&\multicolumn{2}{c}{Case 3}&\multicolumn{2}{c}{Case 4}&\multicolumn{2}{c}{Case 5} \\ 
 				
 				\cmidrule{3-12}
 				
 				~&~& PSNR & SSIM & PSNR & SSIM & PSNR & SSIM& PSNR & SSIM& PSNR & SSIM \\ 
 				 				\midrule
\multirow{7}{*}{\tabincell{c}{HSIs\vspace{0.1cm}\\ {\it WDC}\\{\it PaviaC}\\{\it PaviaU}\\{\it Indian}}} &LRTDTV&{30.88}&{0.888}&{29.55}&{0.849}&{28.29}&{0.825}&{29.35}&{0.843}&{28.31}&{0.824}\\

~&SSTV-LRTF&{30.77}&{0.887}&{30.35}&{0.879}&{28.32}&{0.839}&{29.51}&{0.859}&{27.42}&{0.812}\\

~&HSID-CNN&{29.61}&{0.863}&{22.89}&{0.691}&{21.98}&{0.661}&{22.16}&{0.669}&{21.22}&{0.635}\\

~&SDeCNN&{30.26}&{0.873}&{23.97}&{0.735}&{23.33}&{0.725}&{23.41}&{0.723}&{22.63}&{0.714}\\

~&RCTV&{29.53}&{0.853}&{29.05}&{0.839}&{26.75}&{0.782}&{28.47}&{0.826}&{26.34}&{0.772}\\

~&HLRTF&{30.12}&{0.868}&{29.65}&{0.855}&{29.58}&{0.853}&{29.15}&{0.846}&{29.03}&{0.841}\\

~&H2TF&\bf{32.51}&\bf{0.919}&\bf{31.41}&\bf{0.900}&\bf{31.34}&\bf{0.899}&\bf{30.83}&\bf{0.893}&\bf{30.84}&\bf{0.892}\\
 				\midrule
\multirow{7}{*}{\tabincell{c}{MSIs\vspace{0.1cm}\\ {\it Beads}\\{\it Cloth}\\{\it Cups}}} &LRTDTV&{28.85}&{0.889}&{27.05}&{0.838}&{26.31}&{0.828}&{26.83}&{0.830}&{26.13}&{0.819}\\

~&SSTV-LRTF&{27.64}&{0.878}&{27.48}&{0.864}&{26.25}&{0.855}&{26.79}&{0.844}&{25.11}&{0.823}\\

~&HSID-CNN&{25.86}&{0.827}&{21.22}&{0.660}&{20.97}&{0.645}&{20.68}&{0.646}&{20.34}&{0.626}\\

~&SDeCNN&{28.43}&{0.886}&{22.04}&{0.715}&{22.32}&{0.709}&{21.53}&{0.706}&{21.70}&{0.698}\\

~&RCTV&{28.15}&{0.869}&{27.49}&{0.866}&{25.77}&{0.839}&{26.98}&{0.854}&{25.46}&{0.829}\\

~&HLRTF&{29.21}&{0.884}&{28.73}&{0.886}&{28.67}&{0.884}&{28.10}&{0.870}&{28.03}&{0.868}\\

~&H2TF&\bf{31.51}&\bf{0.940}&\bf{29.46}&\bf{0.906}&\bf{29.47}&\bf{0.901}&\bf{29.22}&\bf{0.908}&\bf{29.03}&\bf{0.896}\\
 				
 				\toprule
 			\end{tabular}
 		}
 	\end{spacing}
 	\vspace{-0.4cm}
 	\label{TABLE:2}
 \end{table}
 \vspace{-0.2cm}
	\section{Experiments}\label{Sec_exp}
	\subsection{Experimental Settings}
We compare H2TF with SOTA model-based methods LRTDTV \cite{lrtdtv}, SSTV-LRTF \cite{SSTVLRTF}, RCTV \cite{RCTV}, and HLRTF \cite{HLRTF} and deep learning methods HSID-CNN \cite{HSID-CNN} and SDeCNN \cite{sdecnn}. We use the pre-trained models of HSID-CNN and SDeCNN provided by authors. All hyperparameters of these methods are carefully adjusted based on authors' suggestions to achieve the best results. We report the peak-signal-to-noise-ratio (PSNR) and structural similarity (SSIM). For more implementation details, please refer to supplementary materials.\par%LRTDTV, SSTV-LRTF, SDeCNN, NGMeet, and RCTV are implemented on Matlab (R2021b). HLRTF and our H2TF are implemented on Pytorch 1.9.0. All experiments are conducted on a server with two Intel(R) Xeon(R) Silver 4210R CPUs, two NVIDIA RTX 3090 GPUs, and 128 GB memory. 
%There are several hyperparameters in our method. The layer numbers of HMF (i.e., $l$) and HNT (i.e., $m$) are set to 5 and 2, respectively. The nonlinear scalar function $\sigma(\cdot)$ is set as the LeakyReLU. The trade-off parameters $\alpha_1$, $\alpha_2$, and $\alpha_3$ are selected in the candidate sets $\{0.4,0.04\}$, $\{0.05,0.5\}$, and $\{0.1,1\}$, respectively, to obtain the best results. The penalty parameter $\mu$ is selected in $\{0.07,0.09,0.11,\cdots, 0.19\}$ to obtain the best results. The parameters $\{r_d\}_{d=1}^{4}$ (the sizes of factor tensors $\{\mathcal{W}_d\}_{d=1}^l$) are selected in $\{(8,16,32,64), (7,14,28,56), (6,12,24,48)\}$ to obtain the best results. The learning rate of the Adam optimizer is set to 0.005. {Since the mask of deadlines (the values of deadlines are zeros) is ready to use, we can replace the fidelity term $\lVert\mathcal{Y}-\mathcal{X}-\mathcal{S}\lVert^2_F$ by $\lVert\mathcal{M}\odot(\mathcal{Y}-\mathcal{X})-\mathcal{S}\lVert^2_F$ for cases with deadlines}, where $\odot$ is the element-wise product and $\mathcal{M}$ is the mask of deadlines.\par
We include four HSIs and three multi-spectral images (MSIs) as simulated datasets. The HSIs are {\it WDC} ($256 \times 256 \times 32 $), {\it PaviaC} ($256 \times 256 \times 32$), {\it PaviaU} ($256 \times 256 \times 32$), and {\it Indian} ($145 \times 145 \times 32$). The MSIs are {\it Beads} ($ 256 \times 256 \times 31 $), {\it Cloth} ($ 256 \times 256 \times 31 $), and {\it Cups} ($ 256 \times 256 \times 31 $) in the CAVE dataset \cite{CAVE_0293}. The noise settings of simulated data are explained as below. {\bf Case 1}: All bands are added with Gaussian noise of standard deviation 0.2. {\bf Case 2}: The Gaussian noise for Case 1 is kept. Besides, all bands are added with impulse noise with sampling rate 0.1. {\bf Case 3}: The same as Case 2 plus 50\% of bands corrupted by deadlines. The number of deadlines for each chosen band is generated randomly from 6 to 10, and their spatial width is chosen randomly from 1 to 3. {\bf Case 4}: The same as Case 2 plus 40\% of bands corrupted by stripes. The number of stripes in each corrupted band is chosen randomly from 6 to 15. {\bf Case 5}: The same as Case 2 plus both the deadlines in Case 3 and the stripes in Case 4. To test our method in real scenarios, we choose two real-world noisy HSIs {\it Shanghai} ($300 \times 300 \times 32$) and {\it Urban} ($307 \times 307 \times 32$) as real-world experimental datasets.
	\begin{figure*}[ht]
	\tiny
	\setlength{\tabcolsep}{0.9pt}
	\begin{center}
		\begin{tabular}{cccccccc}
			\includegraphics [width=0.09\textwidth]{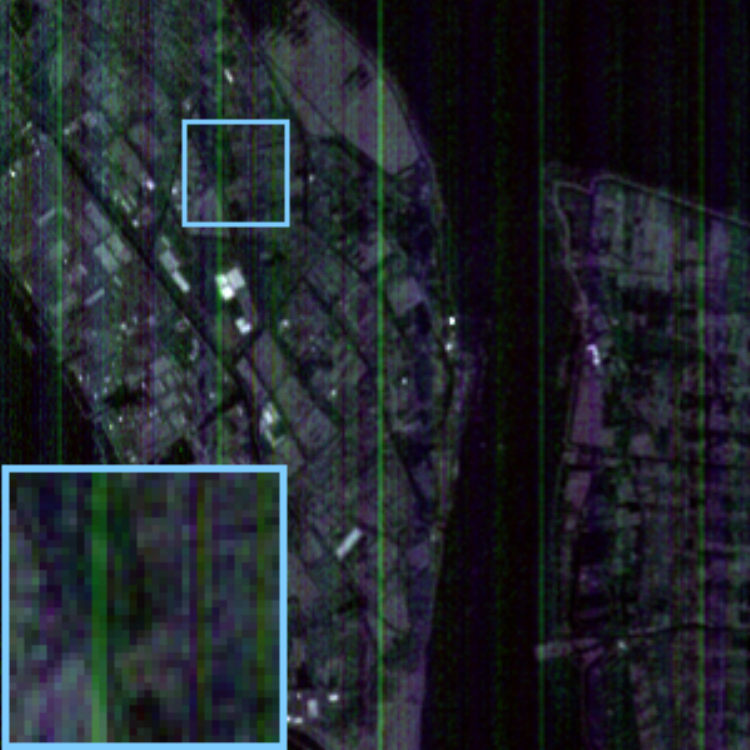}&
			\includegraphics [width=0.09\textwidth]{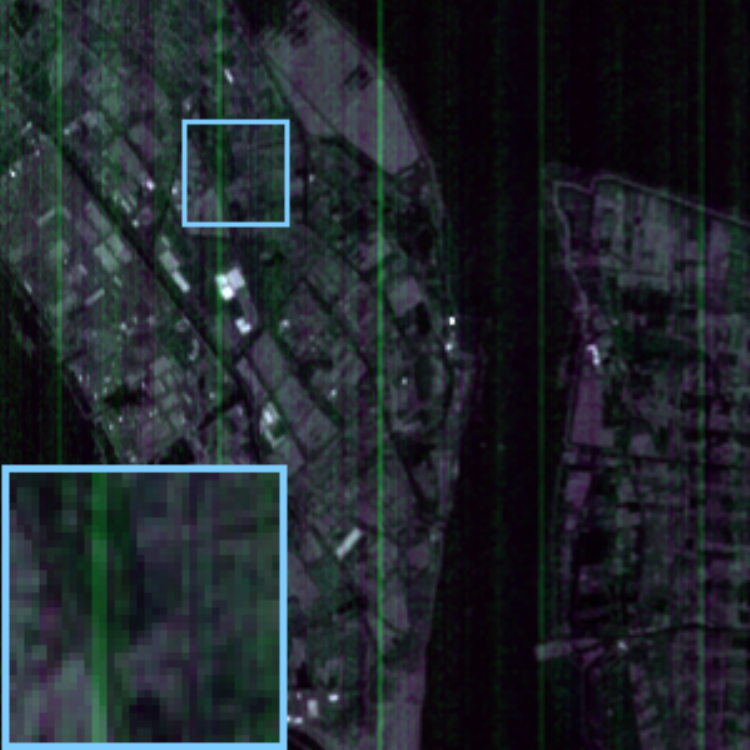}&
			\includegraphics [width=0.09\textwidth]{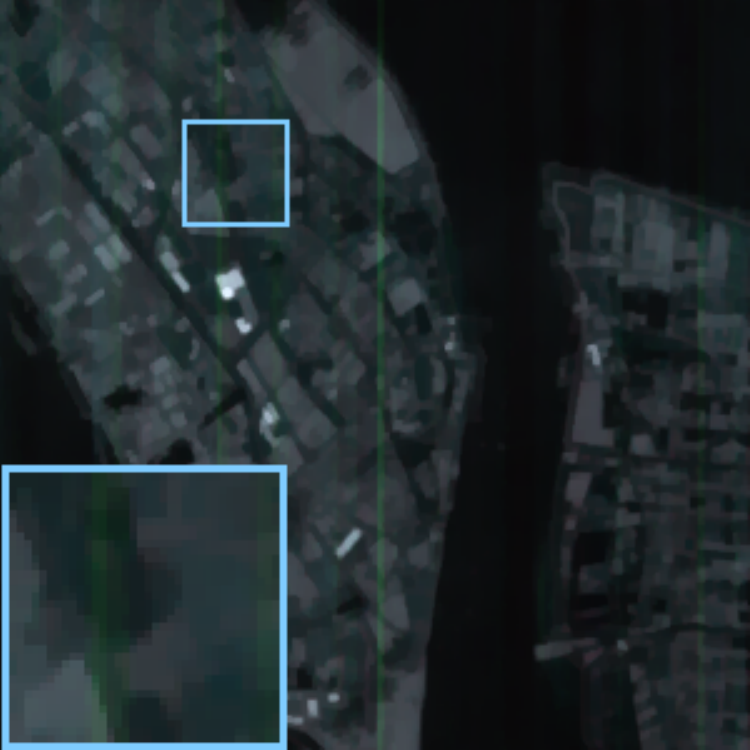}&
			\includegraphics [width=0.09\textwidth]{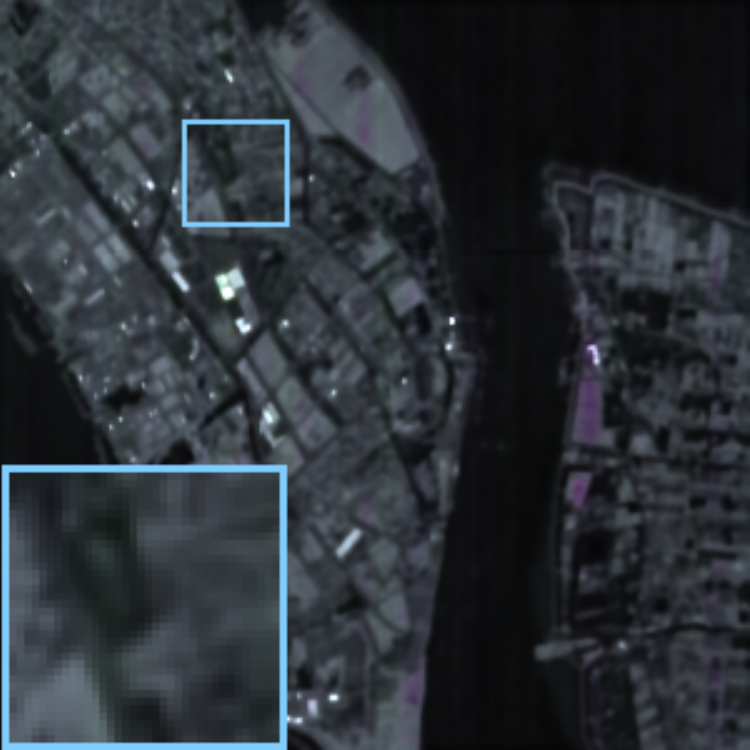}&
			\includegraphics [width=0.09\textwidth]{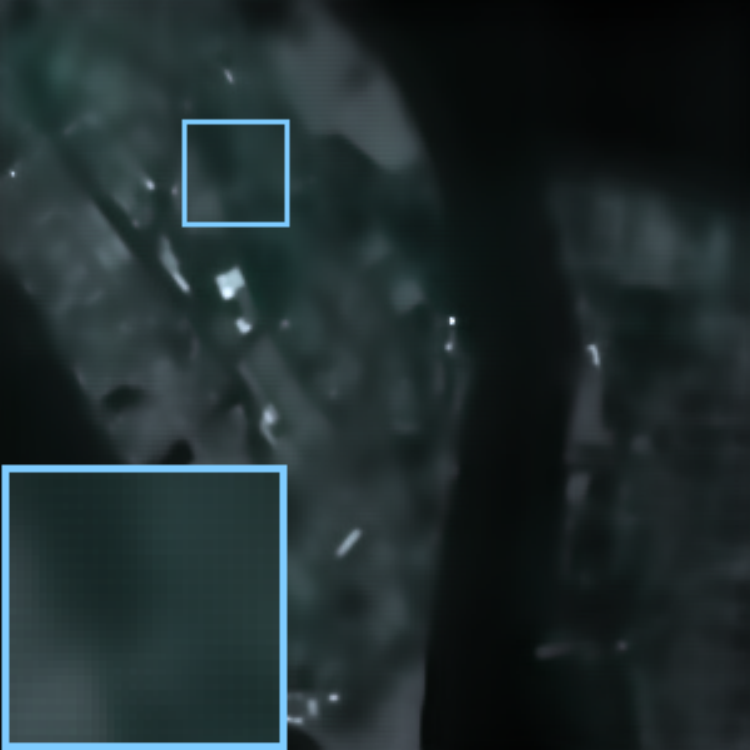}&
			\includegraphics [width=0.09\textwidth]{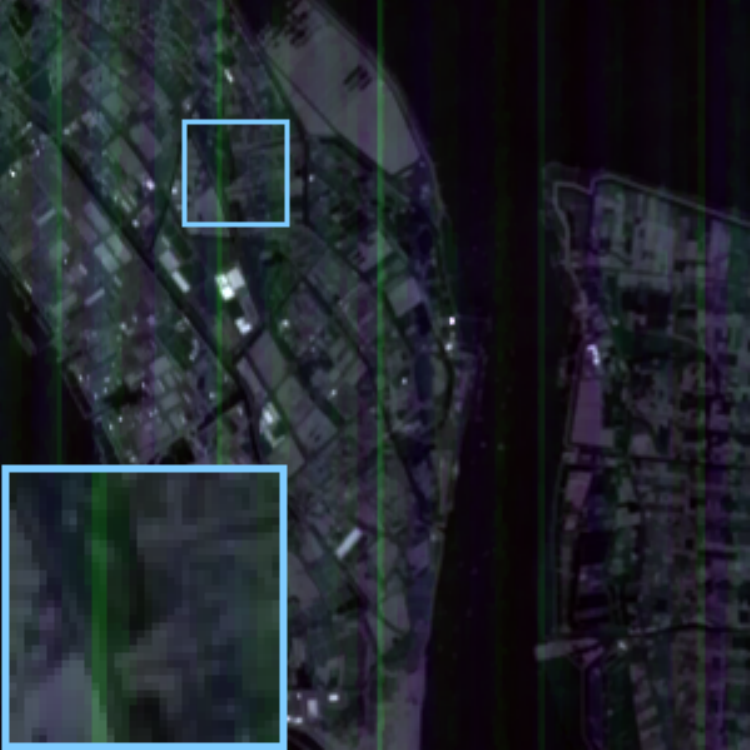}&
			\includegraphics [width=0.09\textwidth]{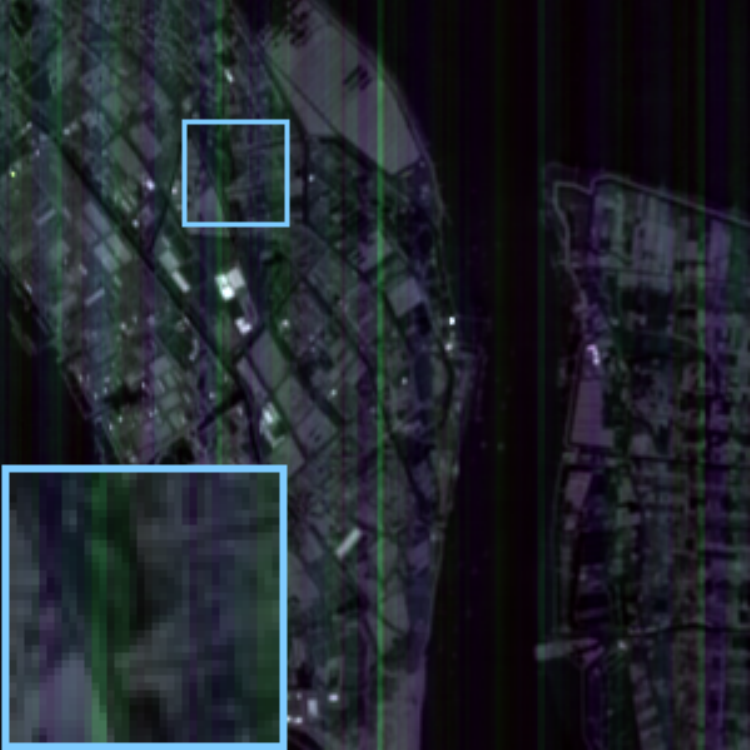}&
			\includegraphics [width=0.09\textwidth]{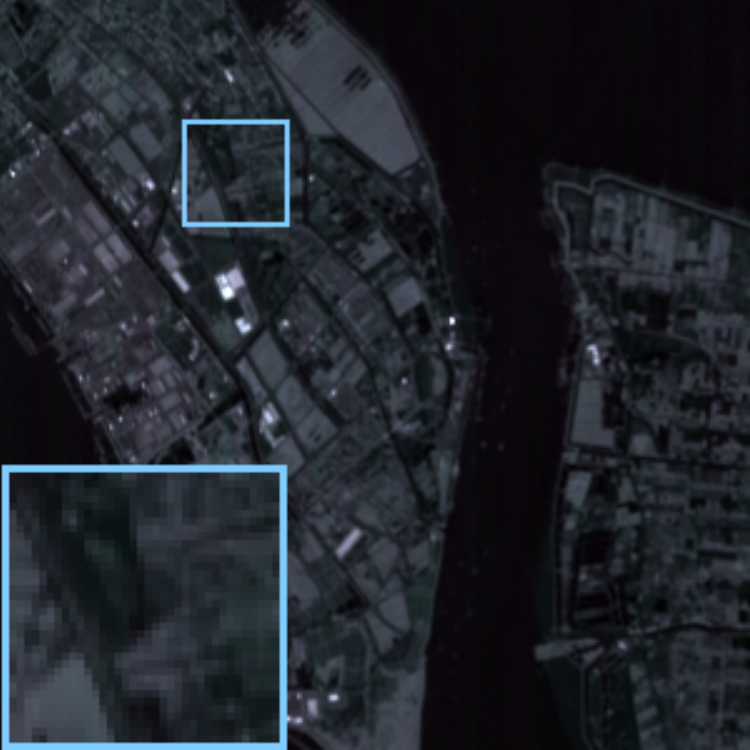}\\
			\includegraphics [width=0.09\textwidth]{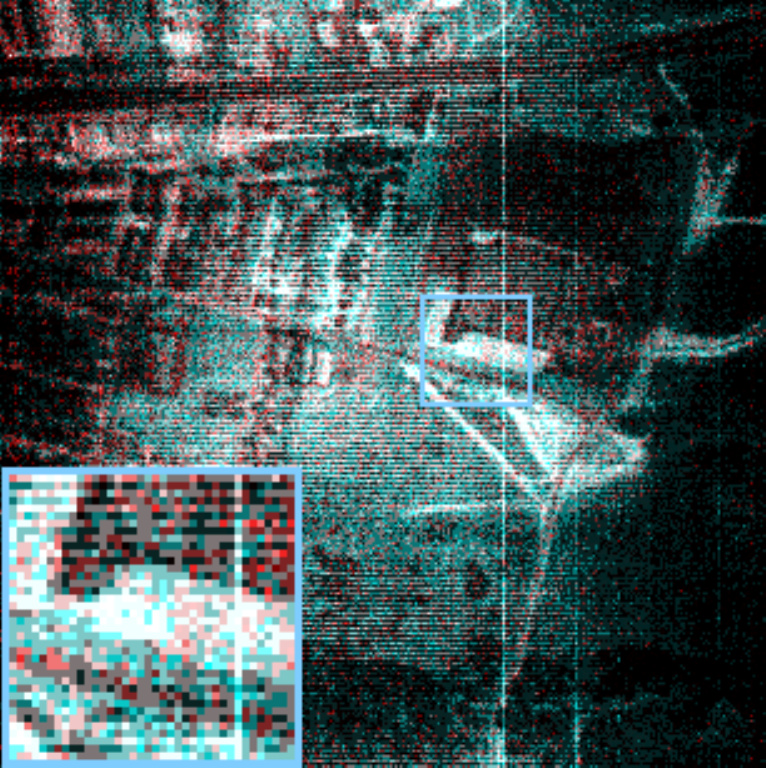}&
			\includegraphics [width=0.09\textwidth]{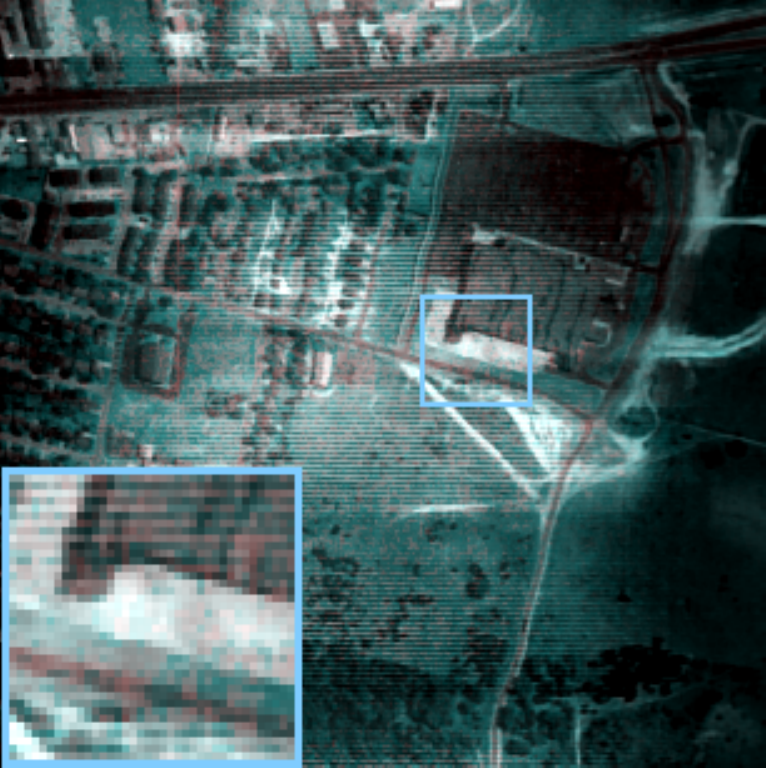}&
			\includegraphics [width=0.09\textwidth]{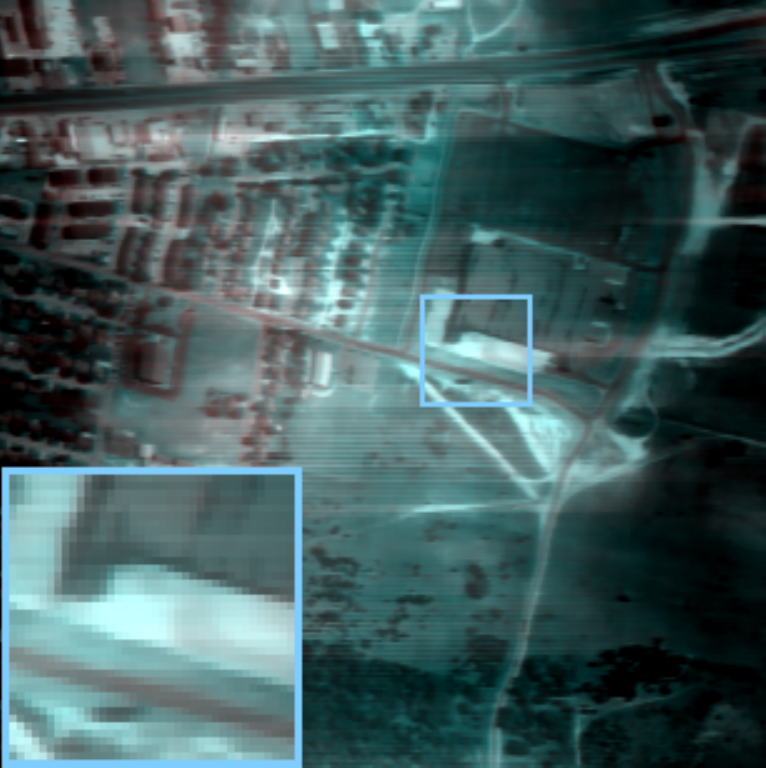}&
			\includegraphics [width=0.09\textwidth]{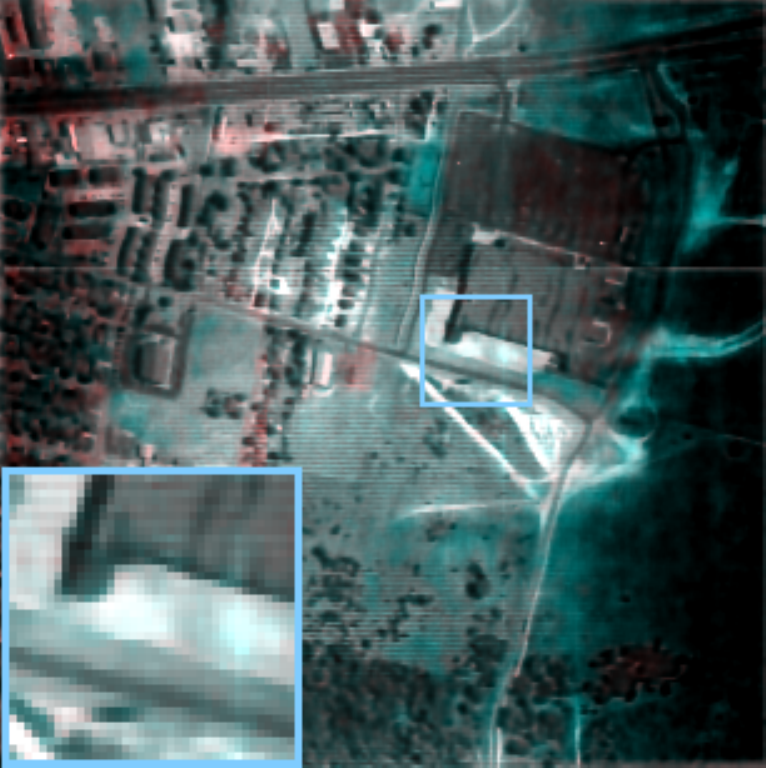}&
			\includegraphics [width=0.09\textwidth]{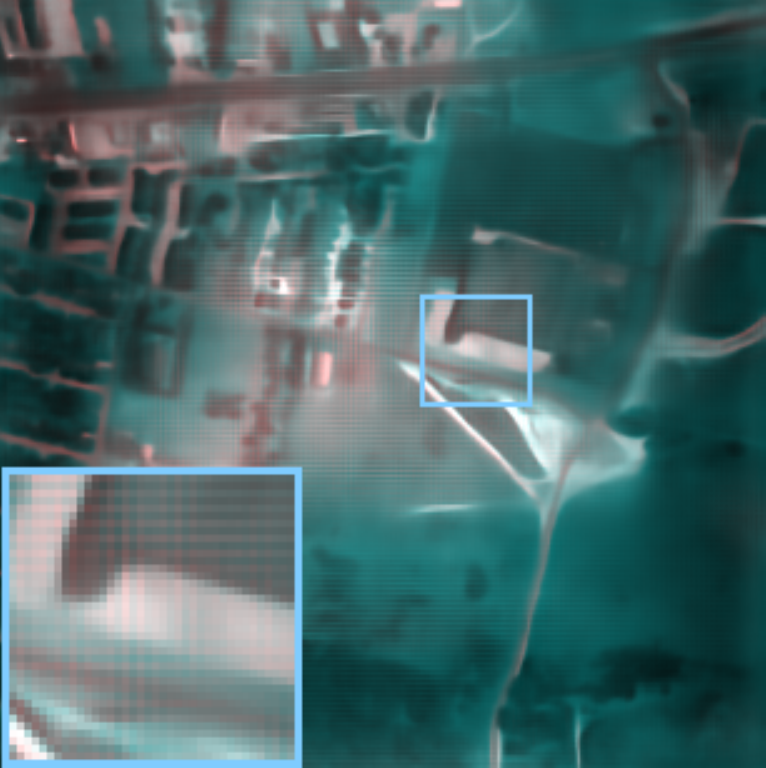}&
			\includegraphics [width=0.09\textwidth]{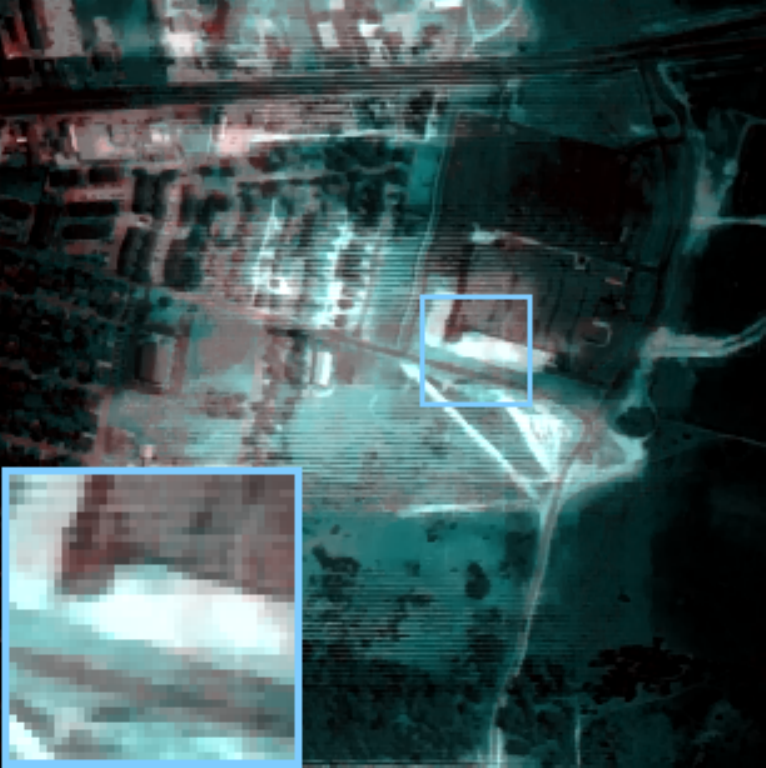}&
			\includegraphics [width=0.09\textwidth]{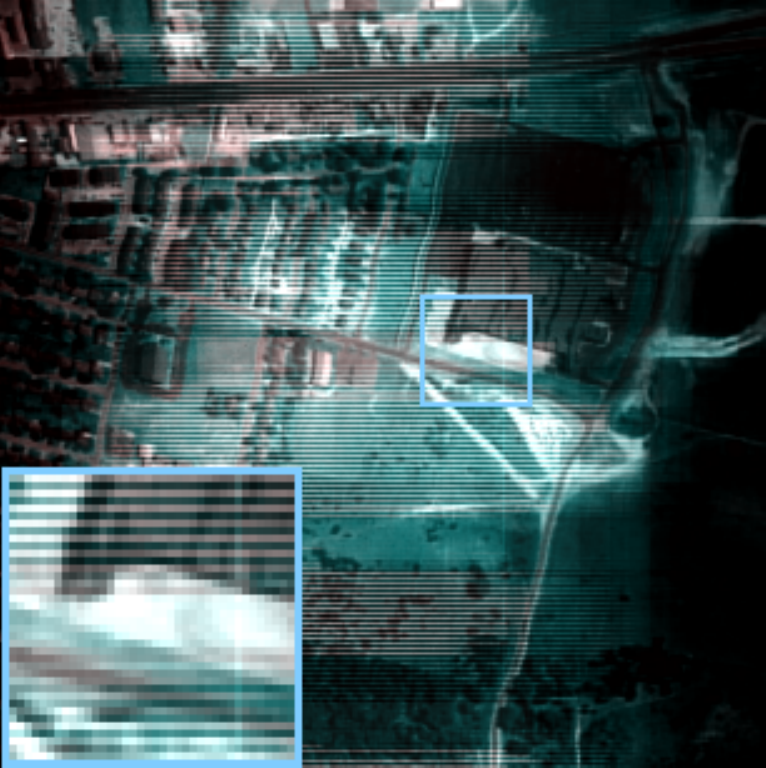}&
			\includegraphics [width=0.09\textwidth]{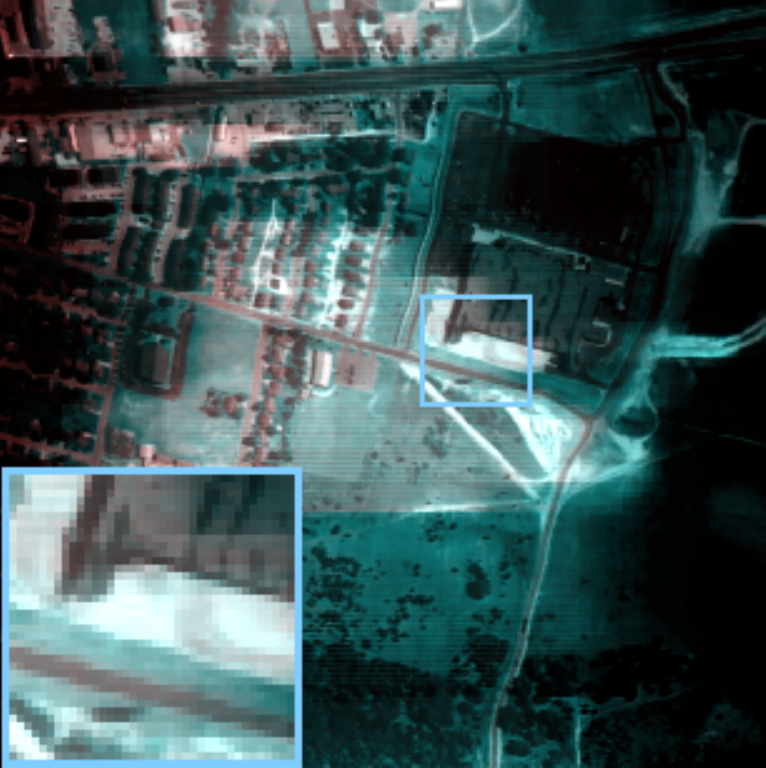}\\
			\vspace{-0.4cm}
			Observed & LRTDTV & SSTV-LRTF & HSID-CNN & SDeCNN & RCTV&HLRTF & H2TF\\
		\end{tabular}
	\end{center}
	\caption{Pseudo-color images of HSI denoising results by different methods on real-world data {\it Shanghai} (first row) and {\it Urban} (second row). }
	\label{figreal}
	\vspace{-0.4cm}
\end{figure*}
\vspace{-0.3cm}
	\subsection{Experimental Results}
	\subsubsection{Results}
	The quantitative results on simulated data are reported in Table \ref{TABLE:2}. Our H2TF obtains better quantitative results than other competitors. H2TF outperforms other TV and tensor factorization-based methods (LRTDTV, SSTV-LRTF, RCTV, and HLRTF), which shows the stronger representation abilities of H2TF than existing shallow tensor factorizations thanks to the hierarchical structures of H2TF. Some visual results on simulated and real data are shown in Figs. \ref{vis}-\ref{figreal}. H2TF generally outperforms other competitors in two aspects. First, H2TF can more effectively remove heavy mixed noise. Second, H2TF preserves fine details of HSIs better than other methods. The superior performances of H2TF are mainly due to its hierarchical modeling abilities, which help to better characterize fine details of HSI and robustly capture the underlying structures of HSI under extremely heavy noise. More visual results can be found in supplementary.  
\begin{figure}[t]
	\centering
	\includegraphics[width=0.48\textwidth]{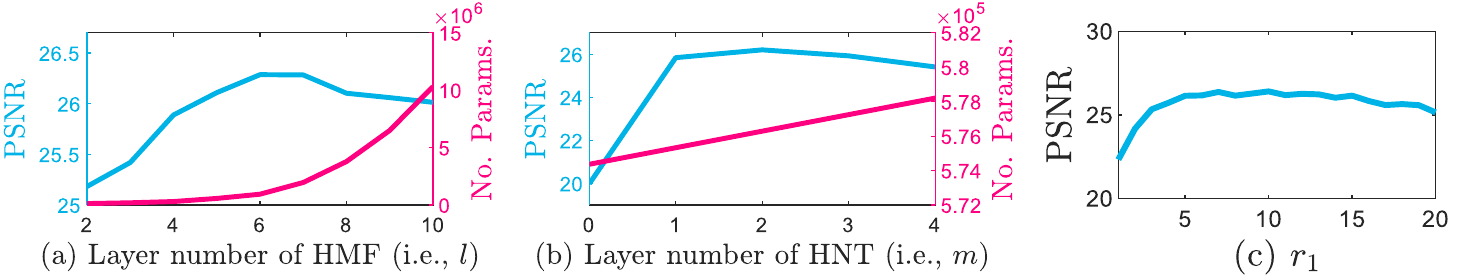}
	\vspace{-0.3cm}
	\caption{Results on {\it Beads} Case 5 with (a) different layer number of HMF, (b) different layer number of HNT, and (c) different sizes of factor tensors.}
	\label{fig:5}
	\vspace{-0.4cm}
\end{figure}
%\begin{figure}[t]
%	\centering
%	\includegraphics[width=0.43\textwidth]{tvs.pdf}
%	\vspace{-0.3cm}
%	\caption{The PSNR value on {\it Beads} Case 5 w.r.t. the iteration number by using our method with or without the sparse regularization $\lVert{\mathcal S}\rVert_{\ell_1}$ and the HSSTV regularization.}
%	\label{regularization}
%	\vspace{-0.3cm} 
%\end{figure}
	\subsubsection{Discussions}
	The HMF is an important building block in H2TF. We test the influence of the layer number of HMF (i.e., $l$); see Fig. \ref{fig:5} (a). A suitable layer number of HMF (e.g., $l=5$) can obtain both good performances and a lightweight model. The HNT is another important building block. We change the layer number of HNT to test its influence; see Fig. \ref{fig:5} (b). Also, a proper layer number of HNT (e.g., $m=2$) can bring good performances. According to Lemma \ref{lemma_1}, {the sizes of factor tensors in HMF, i.e., $\{r_d\}_{d=1}^4$, determine the degree of low-rankness. Hence, we test such connections by changing the sizes of factor tensors; see Fig. \ref{fig:5} (c) (Here, $r_0$ and $r_5$ are fixed as the sizes of observed data and $\{r_d\}_{d=1}^4$ are selected in} $\{(1,2,4,8),(2,4,8,16),$ $(3,6,12,24),\cdots,(20,40,80,160)\}$). When the sizes (rank) are too small, the model lacks representation abilities and when the sizes (rank) are too large, the model overfits. Nevertheless, our method is quite robust w.r.t. $\{r_d\}_{d=1}^4$. %Therefore, our H2TF, which leverages the powerful representation abilities of HMF and HNT, is expected to better characterize fine details of HSIs and obtain more promising results than shallow factorizations.
	%\subsubsection{Effectiveness of HSSTV and Sparse Regularizations}
%We further discuss the effectiveness of HSSTV and sparse regularization $\lVert{\mathcal S}\rVert_{\ell_1}$ in our denoising model (\ref{model_final}). As shown in Fig. \ref{regularization}, the sparsity and HSSTV are both effective and important to ensure the effectiveness of our method for HSI denoising. The results also reveal the compatibility of our H2TF with other hand-crafted regularizations to effectively address the HSI denoising problem. Future work is expected to incorporate H2TF with other regularizations to tackle more multi-dimensional image processing tasks.
\vspace{-0.15cm}
	\section{Conclusions}
	{We propose the H2TF for HSI denoising.} Our H2TF simultaneously leverages the hierarchical matrix factorization and the hierarchical nonlinear transform to compactly represent HSIs with powerful representation abilities, which can more faithfully capture fine details of HSIs than classical tensor factorization methods. Comprehensive experiments validate the superiority of H2TF over SOTA methods, especially for HSI details preserving and heavy noise removal.
	
\bibliographystyle{ieeetr}
\bibliography{ref}
\end{document}